\documentclass[journal]{IEEEtran}
\IEEEoverridecommandlockouts

\usepackage[pdftex]{graphicx}
\usepackage{cite}
\usepackage{amsmath}
\usepackage{array}
\usepackage{url}
\usepackage{xcolor}
\usepackage{tikz}
\usepackage{hyperref}
\usepackage{textcomp}

\usepackage{float}

\newcommand\copyrighttext{%
	\footnotesize \textcopyright 2019 IEEE. Personal use of this material is permitted.
	Permission from IEEE must be obtained for all other uses, in any current or future
	media, including reprinting/republishing this material for advertising or promotional
	purposes, creating new collective works, for resale or redistribution to servers or
	lists, or reuse of any copyrighted component of this work in other works.
	}
\newcommand\copyrightnotice{%
	\begin{tikzpicture}[remember picture,overlay]
	\node[anchor=south,yshift=10pt] at (current page.south) {\fbox{\parbox{\dimexpr\textwidth-\fboxsep-\fboxrule\relax}{\copyrighttext}}};
	\end{tikzpicture}%
}
	
\begin{document}

\title{A Survey of Deep Learning Applications to Autonomous Vehicle Control}

\author{Sampo~Kuutti,~
        Richard~Bowden,~
	Yaochu~Jin,~
	Phil~Barber,~
        and~Saber~Fallah,~
\thanks{This work was supported by the UK-EPSRC grant EP/R512217/1 and Jaguar Land Rover.}%
\thanks{Sampo Kuutti and Saber Fallah are with the Centre for Automotive Engineering, University of Surrey, Guildford, GU2 7XH, U.K. (e-mail: s.j.kuutti@surrey.ac.uk, s.fallah@surrey.ac.uk).}%
\thanks{Richard Bowden is with the Centre for Vision Speech and Signal Processing, University of Surrey, Guildford, GU2 7XH, U.K. (e-mail: r.bowden@surrey.ac.uk).}%
\thanks{Yaochu Jin is with the Department of Computer Science, University of Surrey, Guildford, GU2 7XH, U.K. (e-mail: yaochu.jin@surrey.ac.uk).}%
\thanks{Phil Barber was with Jaguar Land Rover Limited (e-mail: pbarber2@jaguarlandrover.com).}}

\maketitle
\copyrightnotice

\begin{abstract}
Designing a controller for autonomous vehicles capable of providing adequate performance in all driving scenarios is challenging due to the highly complex environment and inability to test the system in the wide variety of scenarios which it may encounter after deployment. However, deep learning methods have shown great promise in not only providing excellent performance for complex and non-linear control problems, but also in generalising previously learned rules to new scenarios. For these reasons, the use of deep learning for vehicle control is becoming increasingly popular. Although important advancements have been achieved in this field, these works have not been fully summarised. This paper surveys a wide range of research works reported in the literature which aim to control a vehicle through deep learning methods. Although there exists overlap between control and perception, the focus of this paper is on vehicle control, rather than the wider perception problem which includes tasks such as semantic segmentation and object detection. The paper identifies the strengths and limitations of available deep learning methods through comparative analysis and discusses the research challenges in terms of computation, architecture selection, goal specification, generalisation, verification and validation, as well as safety. Overall, this survey brings timely and topical information to a rapidly evolving field relevant to intelligent transportation systems.
\end{abstract}

\begin{IEEEkeywords}
Machine learning, Neural networks, Intelligent control, Computer vision, Advanced driver assistance, Autonomous vehicles
\end{IEEEkeywords}

\section{Introduction}
\IEEEPARstart{I}{n} 2016, traffic accidents resulted in 37,000 fatalities in the United States \cite{NationalHighwayTrafficSafetyAdministrationNHTSA2017} and 25,500 fatalities in the European Union \cite{EuropeanCommission2017}. With the steady increase in the number of vehicles on the road, issues such as traffic congestion, pollution, and road safety are becoming critical issues \cite{who2018global}. Autonomous vehicles have gained significant interest as solutions to these challenges \cite{eskandarian2012handbook, Thrun2010, Urmson2008, montanaro2018towards}. For instance, 90\% of all car accidents are estimated to be caused by human errors, while only 2\% are caused by vehicle failures \cite{Singh2015}. Further benefits from autonomous vehicles in terms of better fuel economy \cite{Luettel2012, Payre2014}, reduced pollution, car sharing \cite{Ross2014}, increased productivity, and improved traffic flow \cite{DepartmentforTransport2017} have also been reported.

Some of the earliest autonomous vehicle projects were presented in 1980s by Carnegie Mellon University for driving in structured environments \cite{thorpe1991toward} and the University of Bundeswehr Munich for highway driving \cite{dickmanns1987autonomous}. Since then, projects such as DARPA Grand Challenges \cite{thrun2006stanley, buehler2009darpa} have continued to drive forward research in autonomous vehicles. Outside of academia, car manufacturers and tech companies have also carried out research to develop their own autonomous vehicles. This has led to multiple Advanced Driver Assistance Systems such as Adaptive Cruise Control (ACC), Lane Keeping Assistance, and Lane Departure Warning technologies, which provide modern vehicles with partial autonomy. These technologies not only increase the safety of modern vehicles and make driving easier but also pave the way for fully autonomous vehicles which do not require any human intervention. 

Early autonomous vehicle systems were heavily reliant on accurate sensory data, utilising multi-sensor setups and expensive sensors such as LIDAR to provide accurate environment perception. Control of these autonomous vehicles was handled via rule-based controllers, where the parameters are set by the developers and hand-tuned after simulation and field testing \cite{le2006review, paden2016survey, pasquier2001fuzzylot}. The downside of this approach is the time intensive hand-tuning of parameters \cite{Kuderer2015} and the difficulty of such rule-based controllers to generalise to new scenarios \cite{Silver2013}. Also, the highly non-linear nature of driving means that control methods based on linearisation of the vehicle model or other algebraic analytical solutions are often infeasible or do not scale well \cite{Zhao2013, Desjardins2011}. Recently, deep learning has gained attention due to the numerous state-of-the-art results it has achieved in fields such as image classification and speech recognition \cite{krizhevsky2012imagenet, hinton2012deep, sutskever2014sequence}. This has led to increasing use of deep learning in autonomous vehicle applications, including planning and decision making \cite{schwarting2018planning, mac2016heuristic, veres2011autonomous, caltagirone2017lidar, dixit2018trajectory}, perception \cite{zhu2017overview, van2018autonomous, janai2017computer, benenson2014ten, zhang2016far}, as well as mapping and localisation \cite{lowry2016visual, konda2015learning, kuutti2018survey}. The performance of Convolutional Neural Networks (CNNs) with raw camera inputs has the potential to reduce the number of sensors used by autonomous vehicles. This has led to some organisations investigating autonomous vehicles without expensive sensors such as LIDAR, instead employing extensive use of deep learning for scene understanding, object recognition, semantic segmentation, and motion estimation. The strong results of deep learning in these perception problems have also sparked interest in using Deep Neural Networks (DNNs) to produce control actions in autonomous vehicles. Indeed, autonomous vehicle control often has a strong link to perception, as many techniques use CNNs to predict control actions based on images of the scene, without any separate perception module, thereby removing the separation between the perception and control layer.

Deep learning offers several benefits for vehicle control. The ability to self-optimise its behaviour from data and adapt to new scenarios makes deep learning well suited to control problems in complex and dynamic environments \cite{levine2016end, levine2016learning, Rausch2017}. Rather than having to tune each parameter iteratively while trying to maintain performance in all foreseeable scenarios, deep learning enables developers to describe the desired behaviour and teach the system to perform well and generalise to new environments through learning \cite{mnih2015human, arel2010deep, tani2004self, Lecun2015, schmidhuber2015deep}. For these reasons, there has been significant interest in deep learning for autonomous vehicle control in recent years. There are a variety of different sensor configurations; whilst some researchers aim to control the vehicle with camera vision only, others utilise lower dimensional data from ranging sensors, and some use multi-sensor set ups. There are also some differences in terms of the control objective, some formulate the system as a high-level controller which provides, for example, desired acceleration, which is then realised through a low-level controller, often using classical control techniques. Others aim to learn driving end-to-end, mapping observations directly to low-level vehicle control interface commands. Although there has been a large variety of different approaches used to tackle autonomous vehicle control via deep learning, currently there is a lack of analysis and comparison between these different techniques. This manuscript aims to fill this gap in the literature, by reviewing the deep learning approaches to vehicle control and analysing their performance. Furthermore, the manuscript will evaluate the current state of the field, identify the main research challenges, and make recommendations for the direction of future research.

The remainder of this manuscript is structured as follows. Section II provides a brief introduction to deep learning methods and approaches relevant to autonomous vehicles. Section III discusses recent approaches to autonomous vehicle control using deep learning, which is broken into three categories: (A) lateral, (B) longitudinal, and (C) simultaneous lateral and longitudinal control. Section IV presents the main research challenges from the previous section's discussion. Finally, Section V concludes the current state of the field and provides recommendations for the direction of future research.

\section{Review of Deep Learning}
In this section, we briefly introduce the deep learning techniques and approaches related to the works discussed in later sections. A brief summary on learning strategies, datasets, and tools for deep learning in autonomous vehicles is given. Since a full description on all deep learning algorithms used in autonomous vehicles would be out of the scope of this manuscript, we refer the interested reader to the insightful texts on this topic in \cite{GoodfellowIanBengioYoshuaCourville2016, Nielsen2015, Lecun2015, schmidhuber2015deep, Sutton1998, arulkumaran2017deep, li2017deep}.

\subsection{Supervised Learning}
In deep learning, the objective is to update the weights of a deep neural network during training, such that the model learns to represent a useful function for its task. There are numerous learning algorithms available, but most algorithms described in this manuscript can be classified as supervised or reinforcement learning. Supervised learning utilises labelled data, where an expert demonstrates performing the chosen task at hand. Each data point in the set includes an observation-action pair, which the neural network then learns to model. During training, the network approximates its own action for each observation, and compares the error to the labelled action by the expert. The advantage of supervised learning is speed of training convergence and no need to specify how the task should be performed. While the simplicity of the supervised approach is appealing, the approach has some disadvantages. Firstly, during training the network makes predictions on the control action in an offline framework, where the network's predictions do not affect the states seen during training. However, once deployed, the network's actions will affect future states, breaching the i.i.d. assumption made by most learning algorithms \cite{bottou2008tradeoffs, Ross2011, haan2019causal}. This leads to a distribution shift between training and operation, which can lead to the network making mistakes due to the unfamiliar state distributions seen during operation. Secondly, learning a behaviour from demonstration leaves the network susceptible to biases in the data set. For complex tasks, such as autonomous driving, the diversity of the data set should be ensured if the aim is to train a generalisable model which can drive in all different environments \cite{torralba2011unbiased, gupta2018robot}.

\subsection{Reinforcement Learning}
Reinforcement learning enables the model to learn to perform the task through trial and error. Reinforcement learning can be modelled as a Markov decision process, formally described as a tuple (\textit{S, A, P, R}), where \textit{S} denotes the state space, \textit{A} represents the action space of possible actions, \textit{P} denotes the state transition probability model, and \textit{R} represents the reward function. At each time-step the agent observes a set of states $s_t$, takes an action $a_t$ from possible actions \textit{A}, and then the environment transitions according to \textit{P}. The agent then observes a new set of states $s_{t+1}$ and receives a reward $r_t$. The aim of the agent is to learn a policy $\pi(s_t, a_t)$ mapping observations to actions such that the accumulated rewards are maximised. Therefore, the agent can learn from its own actions through interactions with the environment and receives an estimate of its performance through the reward function. The advantage of this approach is that no labelled data sets are required and a behaviour which generalises well to new scenarios can be learned through reinforcement learning. The downside of reinforcement learning is its low sample efficiency \cite{wang2016sample}, which means converging to an optimal policy can be slow, thereby requiring time-intensive simulations or costly real-world training \cite{Sutton1998}.
	
Reinforcement learning algorithms can be divided into three classes: value-based, policy gradient, and actor-critic algorithms \cite{konda2003onactor}. Value-based algorithms (e.g. Q-learning \cite{watkins1992q}) estimate the value function $V(s)$, which represents the value (expected reward) of being in a given state. If the state transition dynamics \textit{P} are known, the policy can choose actions which bring it to states such that the expected rewards are maximised. However, in most reinforcement learning settings the environment model is not known. Therefore, the state-action value or quality function $Q(s, a)$, which estimates the value of a given action in a given state, is used instead. The optimal policy is then found by greedily maximising the state-action value function $Q(s, a)$. The disadvantage of this approach is that there is no guarantee on the optimality of the learned policy \cite{gordon1995stable, tsitsiklis1996feature}. Policy gradient algorithms (e.g REINFORCE \cite{williams1987reinforcement}) do not estimate a value function, but instead parametrise the policy and then update the parameters to maximise the expected rewards. This is done by constructing a loss function and estimating a gradient of the loss function with respect to the network parameters. During training, the network parameters are then updated in the direction of the policy gradient. The main disadvantage of this approach is the high variance in the estimated policy gradients \cite{sutton2000policy, riedmiller2007evaluation, silver2014deterministic}. The third class, actor-critic algorithms (e.g. A3C \cite{mnih2016asynchronous}), are hybrid methods which combine the use of a value function with a parametrised policy function. This creates a trade-off between the disadvantages of the high variance of policy gradients and the bias of value-based methods \cite{arulkumaran2017deep, grondman2012survey, schulman2015high}. Another separating factor between different reinforcement learning algorithms is the type of reward function used. The reward function used can be either sparse or dense. In a sparse reward function, the agent only receives a reward following specific events, such as success or failure in its task. The benefit of this approach is that the success (e.g. reaching a goal location) or failure (e.g. colliding with another object) is easy to define for most tasks. However, this can further exacerbate the sample complexity issue in reinforcement learning, since the agent would only receive a reward relatively rarely, resulting in slow convergence. On the other hand, in a dense reward function the agent is given a reward at every time-step based on the state it is in. This means that the agent receives a continuous learning signal, estimating how useful the chosen actions were in their respective states.

\subsection{Datasets and Tools for Deep Learning}
The rapid progress in the implementation of deep learning systems on autonomous vehicles has led to the availability of diverse deep learning data sets for autonomous driving and perception. Perhaps the most well known data set for autonomous driving is the KITTI benchmark suite \cite{Geiger2013IJRR}, \cite{Geiger2012CVPR}, which includes multiple data sets for evaluation of stereo vision, optical flow, scene flow, simultaneous localisation and mapping, object detection and tracking, road detection and semantic segmentation. Other useful data sets include the Waymo Open \cite{waymo_open_dataset}, Oxford Robotcar \cite{RobotCarDatasetIJRR}, ApolloScape \cite{huang2018apolloscape}, Udacity \cite{Udacity}, ETH Pedestrian \cite{eth_biwi_00534}, and Caltech Pedestrian \cite{dollar2009pedestrian} data sets. For a more complete overview of available autonomous driving data sets, see the survey by Yin \& Berger \cite{yin2017use}. Besides public data sets, there are also a number of other tools available for the development of deep learning in autonomous vehicles. The current leading Artificial Intelligence (AI) platform for autonomous driving is the NVIDIA Drive PX2 \cite{Nvidiadrive}, which provides two Tegra system-on-chips (SoC) and two Pascal graphics processors with dedicated memory and specialised support for DNN calculations. For more diverse tasks, the MobilEye EyeQ5 \cite{MobilEye} provides four fully programmable accelerators, each optimised for a different family of machine learning algorithms. This diversity can be useful in systems where different families of deep learning algorithms have been used. On the other hand, Altera's Cyclone V \cite{Altera} SoC provides a driving solution optimised for sensor fusion. For a more in-depth review of autonomous driving hardware platforms, see the discussion by Liu et al. \cite{liu2017caad}.

\section{Deep Learning Applications to Vehicle Control}
The motion control of a vehicle can be broadly divided into two tasks; lateral motion of the vehicle is controlled by the steering of the vehicle, whilst longitudinal motion is controlled through manipulating the gas and brake pedals of the vehicle. Lateral control systems aim to control the vehicle's position on the lane, as well as carry out other lateral actions such as lane changes or collision avoidance manoeuvres. In the deep learning domain, this is typically achieved by capturing the environment using the images from on-board cameras as the input to the neural network. Longitudinal control manages the acceleration of the vehicle such that it maintains the desirable velocity on the road, keeps a safe distance from the preceding vehicle, and avoids rear-end collisions. While lateral control is typically achieved through vision, the longitudinal control relies on measurements of relative velocity and distance to the preceding/following vehicles. This means that ranging sensors such as RADAR or LIDAR are more commonly used in longitudinal control systems. The majority of the current research projects have chosen to focus on only one of these actions, thereby simplifying the control problem. Moreover, both types of control systems have different challenges and differ in terms of implementation (e.g. sensor setups, test/use cases). For these reasons this section is split into three subsections, with the first two subsections discussing lateral and longitudinal control systems, independently, and the third subsection focusing on techniques which have attempted to combine both longitudinal and lateral control.
 
\subsection{Lateral Control Systems}
One of the earliest applications of artificial neural networks to the vehicle control problem was the Autonomous Land Vehicle in a Neural Network (ALVINN) system by Pomerleau in 1989 which was first described in \cite{Pomerleau1989} and further extended in \cite{Pomerleau1997}. ALVINN utilised a feedforward neural network, with a 30x32-neuron input layer, one hidden layer with four neurons, and a 30-neuron output layer in which each neuron represents a possible discrete steering action. The system used the input from a camera together with the steering commands of the human driver as training data. To increase the amount of data and variety of scenarios available, the author employed data augmentation methods to increase the available training data without recording any additional footage; each image was shifted and rotated, so as to make the vehicle appear to be situated at a different part of the road laterally. Additionally, to avoid bias towards recent inputs (e.g. if a training session ends in a long right hand turn, the system could be biased to turn right more often) a buffering solution was used where previously encountered training patterns were retained in the buffer. The buffer contained 4 patterns of previous data at any time, which were periodically replaced such that the patterns in the buffer had no right or left bias on average. Both the image shifting as well as buffering solutions were shown to significantly improve the system performance. The system was trained on a 150m stretch of road, after which it was tested on a separate stretch of road at speeds ranging from 5 to 55mph allowing steering without intervention for distances of up to 22 miles. The system was shown to be able to remain, on average, 1.6cm distance from the centre of the road compared to that of 4.0cm under human control. This demonstrated that neural networks can learn to steer a vehicle from recorded data. 

The first to suggest reinforcement learning for vehicle steering was the work carried out by Yu \cite{GeningYu1995}. Yu proposed a road following system based on Pomerleau's work utilising reinforcement learning to design a controller. The advantage of which was the ability to learn from previous experiences to drive in new environments and continuously learn and improve its road following ability through online learning. Combining supervised learning and reinforcement learning, Moriarty et al. \cite{Moriarty1998} developed a lane-selection strategy for a highway environment. The results showed that the vehicles with learned controllers managed to maintain speeds close to the desired speed and resulted in less lane-changes. Moreover, the learned control strategy resulted in better traffic flow than manually constructed controllers.

The neural networks utilised in the aforementioned early works are significantly smaller when compared to what is feasible with today's technology \cite{Bojarski2016}. Indeed, while neural networks are hardly new, the research interest and adoption to various applications has exploded in recent years due to increased computing power, especially through parallel graphics processing units (GPUs) which can significantly reduce training time and improve performance. Moreover, the availability of large public data sets and hardware solutions optimised for deep learning have made training and validation of neural network systems easier. Overall, these recent advancements have enabled better performance through more complex systems with vastly increased amounts of training data and episodes.

\begin{figure}[h]
\centering
\includegraphics[width=2.5in]{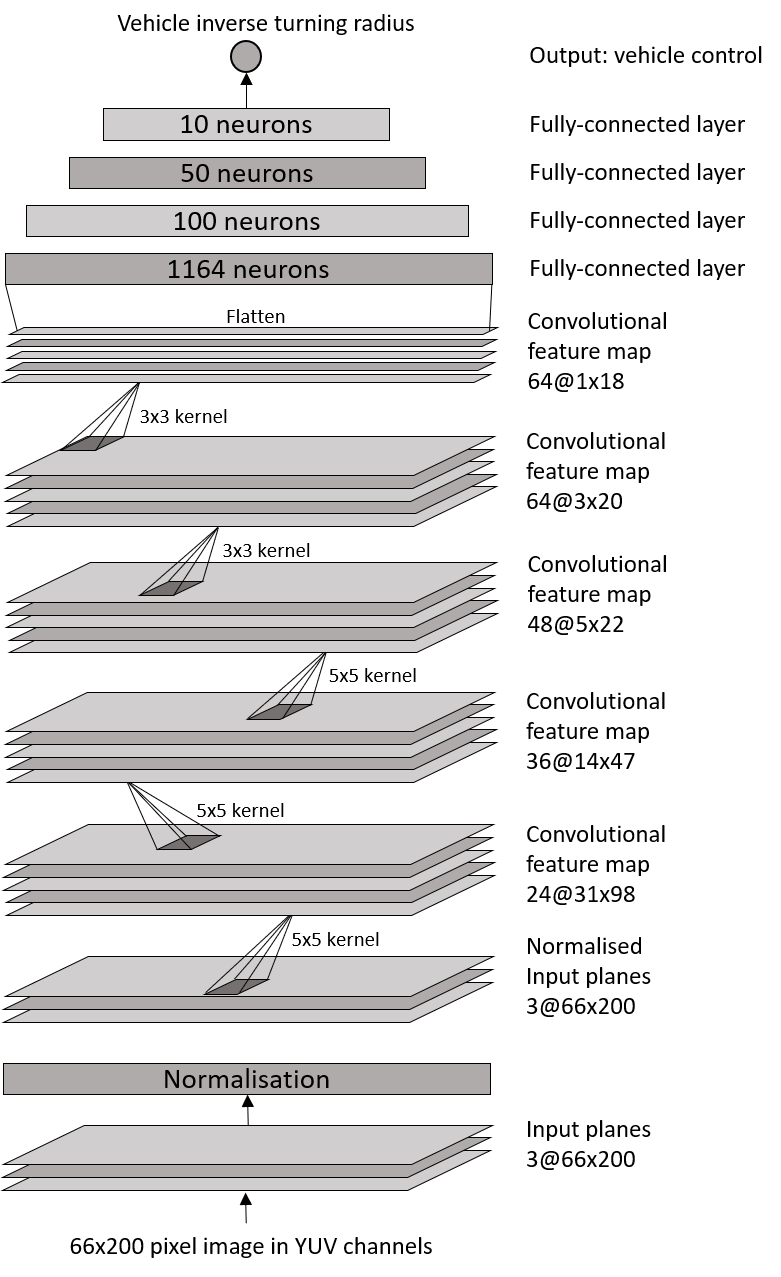}
\caption{Convolutional Neural Network utilised in the NVIDIA end-to-end steering system. (Figure recreated based on \cite{Bojarski2016}).}
\label{fig_nvidiacnn}
\end{figure}

Utilising deeper models with CNNs, Muller et al. \cite{muller2006off} trained a sub-scale radio controlled car to navigate off-road in the DARPA Autonomous VEhicle (DAVE) project. The model was trained with training data collected from two forward-facing cameras while a human was controlling the vehicle. Using a 6-layer CNN, the model learned to navigate around obstacles when driving at speeds of 2m/s. Building on the approach of DAVE, NVIDIA utilised a CNN to create an end-to-end control system for steering of a vehicle through supervised learning \cite{Bojarski2016}. The system is capable of self-optimising the system performance and detecting useful environmental features (e.g. detection of roads and lanes). The CNN used (see Fig. \ref{fig_nvidiacnn}) can learn the steering policy without explicit manual decomposition of the environmental features, path planning, or control actions using a small amount of training data. The training data set consisted of recorded camera footage and steering signals from a human driven vehicle. The CNN consisted of 9 layers, including a normalisation layer, 5 convolutional layers and 3 fully connected layers, with a total of 27 million connections and 250,000 parameters. This method achieved a 98\% autonomy in initial testing and 100\% autonomy during a 10-mile highway test, measured based on the number of interventions required over a given test time. However, it should be noted that this measure does not include lane changes or turns, and therefore only evaluates the system's ability to stay in its current lane.

A further example of supervised learning for steering of an autonomous vehicle is the work by Rausch et al. \cite{Rausch2017}, where supervised learning was employed to create an end-to-end lateral vehicle controller. Rausch et al. utilised a CNN with four hidden layers, three convolutional layers and one fully connected layer. The training data was the steering angle and front-facing camera footage which was provided by a human steering a vehicle in a CarSim \cite{Carsim} simulation, with imaging captured at 12 frames per second (FPS) at a resolution of 1912x1036. The data collection was collected from a 15-minute simulation run resulting in a total of 10,800 frames. Inappropriate frames caused by bad driving behaviour or graphic errors (e.g. due to a fault in the simulator) were removed from the training data manually. Then, the neural network was trained with three different optimisation algorithms to update the network weights, namely Stochastic Gradient Descent (SGD) \cite{bottou2010large}, Adam \cite{kingma2014adam}, and Nesterov's Accelerated Gradient (NAG) \cite{su2014differential}. During training Adam resulted in the best loss convergence, while during the evaluation, the NAG trained network performed the best in terms of keeping the vehicle in the centre of the lane. Therefore, convergence of the loss function is not necessarily representative of a well-trained neural network. The neural networks were shown to learn good estimations of the human driver's steering policy, however by comparing the steering angles, it could be seen that the steering signal of the neural networks included noisy behaviour. A potential reason is that the system estimates the required steering angle at each frame, with no context regarding previous states or actions. This results in the steering signals between subsequent time steps varying significantly from each other, causing noisy output. This could be resolved by utilising a RNN to provide memory of previous inputs and outputs for the system, giving it temporal context.

Introducing temporal context to a deep learning steering model, Eraqi et al. \cite{eraqi2017end} utilised a Convolutional Long Short-Term Memory Recurrent Neural Network (C-LSTM) to learn to steer a vehicle based on visual and dynamic temporal dependencies. The network was trained to predict steering angles based on image inputs, and then compared it to a simple CNN architecture used in \cite{rothe2015dex}. Experimental results showed improved accuracy and smoother steering variations when using the C-LSTM network. However, the model was only evaluated offline by comparing the predicted control action against ground truth, which does not necessarily give an accurate evaluation of driving quality \cite{codevilla2018offline}. Live testing, where the model can control the vehicle to test the learned driving behaviour, should be used instead.

There has also been lateral control techniques for lane change manoeuvrers presented. Wang et al. \cite{wang2018reinforcement} used reinforcement learning to train an agent to execute lane change manoeuvrers using a Deep Q-Network (DQN). The network uses host vehicle speed, longitudinal acceleration, position, yaw angle, target lane, lane width and road curvature to provide a continuous value for the desired yaw acceleration. To ensure Q-learning could be used to output continuous action values, a modified Q-learning approach was used to support continuous action values, where the Q-function was a quadratic function approximated by three single hidden layer feedforward neural networks. The proposed approach was tested in a simulated highway environment, with preliminary results showing effective lane change manoeuvrers learned by the agent.

A summary of the research works covered in this section can be seen in Table \ref{table_lat}. Due to the advancements mentioned previously, the recent trend has been to move to deeper models with increased amounts of training data. Recent works have also investigated introducing temporal cues into the learning model, but this suffers from instability in training. Moreover, many of the models developed so far have been trained and evaluated in relatively simple environments. For instance, most researchers have decided to focus on lateral control for a single task. For example in models trained for lane keeping no decision-making for e.g. lane changes or turns to different roads have been incorporated in these systems. This opens possible avenues for future research where multiple actions could be carried out by the same DNN. It should also be noted that the majority of these works were trained and evaluated in simulated environments, which further simplifies the task and would require further tests to validate their real world performance. Nevertheless, there have been important developments in this field and these results show great promise for the use of deep learning for autonomous vehicle control.

\begin{table*}[t]
	\scriptsize
	\renewcommand{\arraystretch}{1.5}
	\caption{A Comparison of Lateral Control Techniques.}
	\label{table_lat}
	\centering

	\begin{tabular}{ | m{1.5em} | m{2cm} | m{2cm} | m{2cm} | m{2cm} | m{2cm} | m{2cm} | m{2cm} |}
	\hline
	Ref. & Learning Strategy & Network & Inputs & Outputs & Pros & Cons & Experiments \\
	\hline
	\cite{Pomerleau1989}, \cite{Pomerleau1997} & Supervised Learning & Feedforward network with 1 hidden layer & Camera image & Discretised steering angles & First promising results for neural network-based vehicle controllers & Simple network and discretised steering angle outputs degrade performance & Real \& Simulation \\
	\hline
	\cite{GeningYu1995} & Reinforcement Learning & Feedforward network with 1 hidden layer & Camera image & Discretised steering angles & Supports online learning & Simple network and discretised steering angle outputs degrade performance & Simulation \\
	\hline
	\cite{muller2006off} & Supervised Learning & 6-layer CNN & Camera images & Steering angle & Robust to environmental diversity & Large errors, trained and tested on a sub-scale vehicle model & Real world (sub-scale vehicle) \\
	\hline
	\cite{Bojarski2016} & Supervised Learning & 9-layer CNN & Camera image & Steering angle values & High level of autonomy during field tests & Only considers lane following, requires interventions by the driver & Real world \& Simulated \\
	\hline
	\cite{Rausch2017} & Supervised Learning & 8-layer CNN & Camera image & Steering angle values & Learns from minimal training data & Noisy behaviour of the steering signal & Simulation \\
	\hline
	\cite{eraqi2017end} & Supervised Learning & C-LSTM & Camera image & Steering angle values & Considers temporal dependencies & RNNs can be difficult to train, lack of live testing & No live testing, tested on data set image examples only \\
	\hline
	\cite{wang2018reinforcement} & Reinforcement Learning & 3 feedforward networks & Host vehicle states and road geometry & Vehicle yaw acceleration & Executes lane changes successfully & Limited testing or results, lack of comparison to other lane change algorithms & Simulation \\
	\hline
	\end{tabular}
\end{table*}

\subsection{Longitudinal Control Systems}
Machine learning methods have also shown promise in applications to vehicle longitudinal control, such as ACC design. The ACC can be described as an optimal tracking control problem for a complex nonlinear system \cite{vahidi2003research, moon2009design} and therefore is poorly suited to control systems based on linear vehicle models or other algebraic analytical solutions \cite{Sun2016}. Such traditional control systems provide poor adaptability in complex environments and do not conform to the driver's habits \cite{Chen2017}. The strong nonlinear nature of the system makes it difficult to build a vehicle model without significant uncertainty, limiting the effectiveness of model-based solutions. However, neural networks have shown great potential for optimising nonlinear, high-dimensional control systems \cite{levine2016end, levine2016learning, wang2005neural, polycarpou1996stable, sanner1992stable, wang2002adaptive, ren2009adaptive, zhang2000adaptive}. For instance, reinforcement learning can learn an optimal control policy through interaction with the environment, without knowledge of the system model \cite{Sutton1998}. Furthermore, the strong adaptive capacity and model-free capability of reinforcement learning makes it an attractive solution for ACC design. In early works, Dai et al. \cite{Dai2005} proposed a fuzzy reinforcement learning method for longitudinal control of an autonomous vehicle. The method combines a Q estimator network (QEN) with a Takagi-Sugeno-type Fuzzy Inference System (FIS). The QEN is used to estimate the optimal action value function whilst the FIS gets the control output based on the estimated action value function. The described approach was evaluated in a simulation of a car-following scenario where the lead vehicle varies its velocity over time with a maximum episode duration of 80s. The controller was shown to be able to successfully drive the vehicle without failing after 68 trials. However, the reward function of the proposed approach by Dai et al. is only based on the spacing between the lead and the following vehicle. The reward function is the key to a successful reinforcement learning approach as it is the means by which the developer indicates the desirability of being in any given state. Therefore, the reward function needs to accurately capture the task to be performed and the manner in which it should be completed. For longitudinal control, the reward function should motivate the agent to adopt a safe and efficient driving strategy. For these reasons, a reward function with only one parameter such as inter-vehicle spacing may not be sufficient in real-time applications.

There are several works in which the use of multi-objective reward functions have been explored. For example, Desjardins \& Chaib-Draa \cite{Desjardins2011} used a multi-objective reward function based on time headway (distance in time from the lead vehicle) and time headway derivative. The agent was encouraged through the reward function to keep a 2s time headway to the lead vehicle, and the time headway derivative provided information regarding whether the vehicle is moving closer to or farther from the lead vehicle, and allowed it to adjust its driving strategy accordingly. Taking the time headway derivative into consideration in the reward function encourages the agent to choose actions which help it progress toward the desired state (ideal time headway). The authors used this reward function in a policy-gradient method for a Cooperative Adaptive Cruise Control (CACC) system. The neural network architecture chosen had two inputs, a single hidden layer of 20 neurons, and an output layer with 3 discrete actions (brake, accelerate, do nothing). In the learning process, an average of over 2.2 million iterations were obtained over ten learning simulations. The chosen method was shown to be efficient in CACC, providing average time headway errors of 0.039s in an emergency braking scenario. While the magnitude of the time headway errors remain small, it should be noted that the velocity profile of the subject vehicle showed oscillatory behaviour. This would make the system uncomfortable for the passengers as well as pose a potential safety risk. Potential solutions for this could include utilising continuous action values, the use of RNNs, or negative rewards for changes in acceleration to help smooth the velocity profile of the vehicle. Similarly, Sun \cite{Sun2016} proposed a CACC system based on rewards from time headway and time headway derivative in a Q-learning algorithm. This approach was shown to reduce the learning time of the neural network. Over one hundred learning simulations, the best performing policy (the policy which obtained the highest reward) was chosen for evaluation. The algorithm was evaluated in a simulation of a stop-and-go environment in which the lead vehicle accelerated and decelerated periodically. The agent was shown to provide adequate performance in a platoon scenario. However, whilst such multi-objective reward functions are an improvement over single objective reward functions such as the one proposed by Dai et al. \cite{Dai2005}, this reward function does not consider passenger comfort which could lead to harsh accelerations or decelerations.

Huang et al. \cite{Huang2017} presented a Parameterised Batch Actor-Critic (PBAC) reinforcement learning algorithm for longitudinal control of autonomous vehicles based on actor-critic algorithms. A multi-objective reward function was designed to reward the algorithm for tracking precision and drive smoothness. The method was validated by field experiments on various driving environments (e.g. flat, slippery, sloping, etc.) and the results suggested the method can track time-varying speeds more precisely than traditional Proportion-Integration (PI) or Kernel-based Least Square Policy Iteration (KLSPI) controllers trained with reinforcement learning \cite{xu2007kernel, Wang2014}. This was due to lower sensitivity to noise of speeds and accelerations. Moreover, smooth driving was achieved using the proposed method. The addition of driving smoothness in the reward function makes these systems more comfortable for passengers. However, the method was evaluated in an environment without adjacent vehicles or other obstacles. This allowed the authors to not consider safety parameters  in the reward function, which leaves the algorithm susceptible to crashes in environments with other vehicles present. Therefore, additional terms for safety would be required in the reward function to ensure safe behaviour of the autonomous vehicle. \vskip-1ex

\begin{figure}[h]
	\centering
	\includegraphics[width=3.5in]{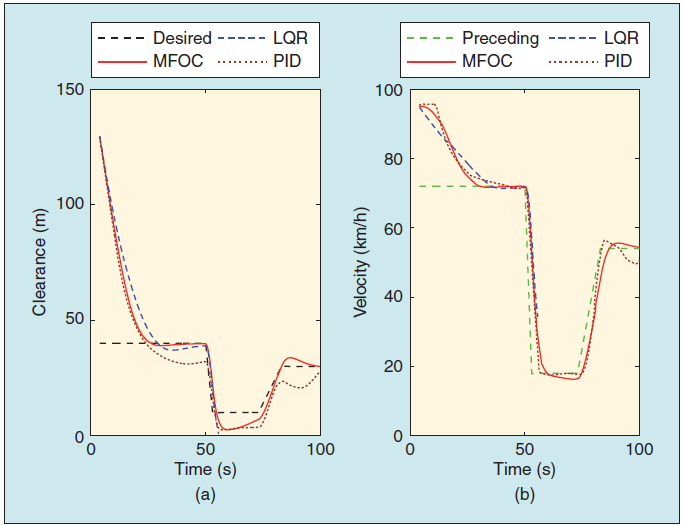}
	\caption{MFOC Controller compared to PID and LQR controllers in an emergency braking scenario. (a) Clearance between the lead and follower vehicle. (b) Velocity profiles of the lead vehicle and the three controllers \cite{Zhao2017}.}
	\label{fig_mfoc}
\end{figure}

One such reward function was proposed by Chae et al. \cite{Chae2017}, who proposed an autonomous braking system for collision avoidance based on a DQN approach. The reward function balances two conflicting objectives: avoiding collision and getting out of high risk situations. To speed up convergence, a replay memory was used to store a number of episodes of which some are chosen randomly to help train the network. Additionally, a 'trauma memory' of rare critical events (e.g. collision) was used to improve stability and make the agent more reliable. The system was evaluated in situations where the vehicle had to avoid collision with a pedestrian, using various Time-to-Collision (TTC) values with 10,000 tests for each TTC value. It was shown that for TTC values above 1.5s, collisions were avoided every time, whereas at 0.9s (lowest TTC value used) the collision rate was as high as 61.29\%. Additionally, the system was evaluated in a test procedure specified by the Euro NCAP test protocol (CVFA and CVNA tests \cite{euroncap2015}) and the system passed these tests without collision. Therefore, the system was considered to exhibit desirable and consistent brake control behaviour. In addition, Chen et al. \cite{Chen2017} presented a personalised ACC which can learn from human demonstration. The proposed algorithm is based on Q-learning with a reward function based on distance to the front vehicle, vehicle speed, and acceleration. The authors used a Q-learning algorithm based on a feedforward artificial neural network to estimate the Q-function and calculated the desired velocity, which is then converted to low-level control commands by a Proportional Integral Derivative (PID) controller. The neural network used to estimate the Q-function consists of an input layer with 5 nodes, a hidden layer with 3 nodes, and an output layer with 1 node which predicts the desired velocity. The performance of the system was evaluated based on comfort and driving smoothness in simulation with different velocities and desired inter-vehicle clearances. The system was shown to provide better performance when compared to traditional ACC approaches. Similarly, Zhao et al. \cite{Zhao2017} proposed a personalised ACC approach which considers safety, comfort, as well as personalised driving styles. The reward function considers the driver habits, passenger comfort, and safety in an effort to find a good tradeoff between safety and comfort. The proposed approach uses a Model-free Optimal Control (MFOC) algorithm based on an actor-critic neural network structure. By optimising the algorithm to drive in a more human-like fashion, the human driver is more likely to trust the system and continue using it. For this purpose, the network would also be capable of learning from the human driver when the cruise control feature was switched off to better tune its parameters and to adopt a driving strategy based on the owner's driving habits. The proposed algorithm was tested in a simulation under various environments and was shown to perform better than PID and Linear Quadratic Regulator (LQR) based controllers. For instance, in an emergency braking test scenario shown in Fig. \ref{fig_mfoc}, the MFOC maintained a safer clearance compared to PID, while the LQR failed the test by causing a rear end collision. However, while conforming to individual driving habits can be useful to ensure the user feels safe and comfortable in the car, strategies for mitigating the negative effects of learning bad driving habits should also be considered to ensure the long term reliability and safety of the system.

Reinforcement learning has been shown to be an effective approach for vehicle longitudinal control systems as shown by the discussion above. However, the main drawback for reinforcement learning is the time-intensive training \cite{Sutton1998, kaelbling1996reinforcement}. In contrast, supervised learning methods simplify the learning process with the use of prior knowledge of the supervisor, but lack the level of adaption that makes reinforcement learning attractive to complex decision-making systems such as autonomous driving. For these reasons, there are multiple examples in the literature that combine reinforcement and supervised learning to exploit the advantages of both approaches; reinforcement learning allows for self-adaptation in new and complex environments whilst the prior knowledge of supervised learning speeds up the learning process. For example, Zhao et al. \cite{Zhao2013, Zhao2014, Wang2015} introduced a supervised reinforcement learning algorithm for an ACC system. By utilising actor-critic methods, the authors propose a novel supervised actor-critic (SAC) learning scheme, which is then implemented with feed-forward neural networks into a hierarchical acceleration controller. The proposed approach was evaluated in a simulation for an emergency braking scenario. The network was trained for emergency braking in dry conditions, whilst it was evaluated in both dry and wet road conditions and results were compared to the performance of a PID controller. The simulation results demonstrated that the SAC algorithm has superior performance compared to that of the PID controller as well as a supervised learning based controller (without reinforcement learning), and can adapt to changing road conditions. This shows the benefits of combining supervised learning with reinforcement learning to leverage the combined advantages of both methods. Pre-training the network via supervised learning helps reduce the training time of reinforcement learning and improves the convergence of the algorithm, both of which are common problems in reinforcement learning algorithms. Meanwhile, by exploring different actions through trial and error, reinforcement learning improves the performance beyond what supervised learning can provide. Also, the authors stated that using an actor-critic network architecture was beneficial as the evaluation of actions by the critic boosts the system's performance in critical scenarios such as emergency braking.

A summary of the longitudinal control methods can be seen in Table \ref{table_long}. In contrast to lateral control systems, vision-based inputs are not generally used for longitudinal control. Instead sensor inputs from ranging sensors (e.g. RADAR, LIDAR) and host vehicle states are more commonly used. These lower dimensional inputs (e.g. time headway or relative distance) can then easily be used to define a reward function for reinforcement learning. The second major difference between lateral and longitudinal control algorithms is the choice of learning strategies. While lateral control techniques favour supervised learning techniques trained on labelled datasets, longitudinal control techniques favour reinforcement learning methods which learn through interaction with the environment. However, as seen in this section, the reward function in reinforcement learning needs to be carefully designed. Safety, performance, and comfort all need to be considered. Poorly designed reward functions result in poor performance or the model not converging. Another challenge with reinforcement learning algorithms is the trade-off between exploration and exploitation. During training, the agent must take random actions to explore the environment. However, to perform well in its task the agent should exploit its knowledge to find the optimal action. Example solutions for this are the $\epsilon$-greedy exploration policies and the Upper Confidence Bound (UCB) algorithm. $\epsilon$-greedy strategies choose a random action with a probability $\epsilon$, which decreases overtime as the agent learns its environment. On the other hand, UCB encourages exploration in states with high uncertainty, whilst exploitation is encouraged in regions with high confidence. Therefore, intrinsic motivation is implemented in the system, encouraging the agent to learn about its environment, whilst exploitation can be taken advantage of in states which have already been explored adequately \cite{arulkumaran2017deep, lai1985asymptotically, bellemare2016unifying, schmidhuber1991possibility}. Other approaches have sought to use supervised learning as a pre-training step to get the advantages of both reinforcement and supervised learning. 

\begin{table*}[t]
	\scriptsize
	\renewcommand{\arraystretch}{1.2}
	\caption{A Comparison of Longitudinal Control Techniques.}
	\label{table_long}
	\centering
	
	\begin{tabular}{ | m{1.5em} | m{2cm} | m{2cm} | m{2cm} | m{2cm} | m{2cm} | m{2cm} | m{2cm} |}
		\hline
		Ref. & Learning Strategy & Network & Inputs & Outputs & Pros & Cons & Experiments\\
		\hline
		\cite{Dai2005} & Fuzzy Reinforcement Learning & Feedforward network with 1 hidden layer & Relative distance, relative speed, previous control input & Throttle angle, brake torque & model-free, continuous action values & Single term reward function & Simulation \\
		\hline
		\cite{Desjardins2011} & Reinforcement Learning & Feedforward network with 1 hidden layer & Time headway, headway derivative & Accelerate, brake, or no-op & Maintains a safe distance & Oscillatory acceleration behaviour, no term for comfort in reward function & Simulation \\
		\hline
		\cite{Huang2017} & Reinforcement Learning & Actor-Critic Network with feedforward networks & Velocity, velocity tracking error, acceleration error, expected acceleration & Gas and brake commands & Learns from minimal training data & Noisy behaviour of the acceleration signal & Real world \\
		\hline
		\cite{Chae2017} & Reinforcement Learning & Feedforward network with 5 hidden layers & Vehicle velocity, relative position of the pedestrian for past 5 time steps & Discretised deceleration actions & Reliably avoids collisions & Only considers collision avoidance with pedestrians, high rate of collision at low TTC & Simulation \\
		\hline
		\cite{Chen2017} & Reinforcement Learning & Feedforward network with 1 hidden layer & Relative distance, relative velocity, relative acceleration (normalised) & Desired acceleration & Provides smooth driving styles, learns personal driving styles & No methods for preventing learning of bad habits from human drivers & Simulation \\
		\hline
		\cite{Zhao2017} & Reinforcement Learning & Actor Critic Network with feedforward networks & Relative distance, host velocity, relative velocity, host acceleration & Desired acceleration & Performs well in a variety of scenarios, safety and comfort considered, learns personal driving styles & Adapting unsafe driver habits could degrade safety & Simulation \\
		\hline
		\cite{Zhao2013} & Supervised Reinforcement Learning & Actor-Critic Network with feedforward networks & Relative distance, relative velocity & Desired acceleration & Pre-training by supervised learning accelerates learning process and helps guarantee convergence, performs well in critical scenarios & Requires supervision to converge, driving comfort not considered & Simulation \\
		\hline
	\end{tabular}
\end{table*}

\subsection{Simultaneous Lateral \& Longitudinal Control Systems}
\begin{figure}[h]
	\centering
	\includegraphics[width=3.5in]{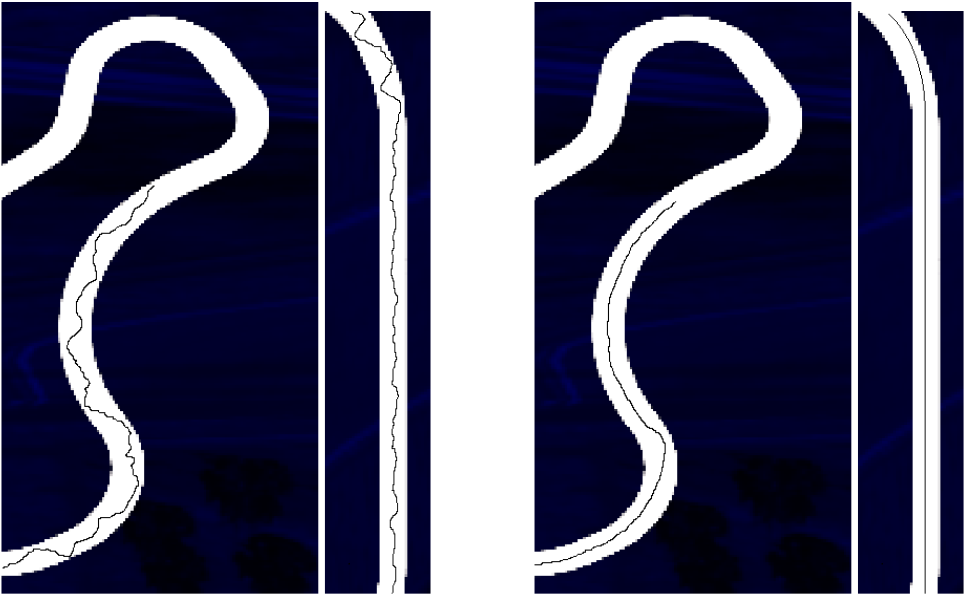}
	\caption{The lane keeping performance of (a) the DQN with discretised outputs and (b) DDAC with continuous output values \cite{Sallab2016}.}
	\label{fig_dqnvsddac}
\end{figure}

The previous sections demonstrated that DNNs can be trained for either longitudinal or lateral control of a vehicle. However, for autonomous driving, the vehicle must be able to control both steering and acceleration simultaneously. In early works towards full vehicle control through deep learning, Xia et al. \cite{WeiXia12016} introduced an autonomous driving system based on Q-learning combined with learning from the experience of a professional driver. The reward value of the professional driver's strategy and the Q-value learned through the Q-learning method were combined in the pre-training phase to improve the speed of convergence during training. A filtered experience replay stores a limited number of episodes and allows elimination of poor experimental rounds from memory, improving convergence on a control strategy. The proposed Deep Q-learning with filtered experiences (DQFE) approach was compared to a naive neural fitted Q-iteration (NFQ) \cite{riedmiller2005neural} algorithm without pre-training by an experienced driver. During training, it was shown that the DQFE approach reduced the training time by 71.2\% for the 300 training episodes. Moreover, during 50 tests on a competition track, the proposed approach completed the track 49 times, compared to only 33 with NFQ. Additionally, DQFE performed better in terms of mean distance from centre of the track. Therefore, the addition of filtered experience replay improved the speed of convergence as well as performance of the algorithm. Comparing two neural networks for lane keeping systems, Sallab et al. \cite{Sallab2016} investigated the effects of discretised and continuous actions. Two approaches, DQN and a Deep Deterministic Actor Critic (DDAC) algorithm, were evaluated in a TORCS simulator \cite{TORCS}. In the two networks developed by the authors, the DQN could only output discretised values (steer, gear, brake, and acceleration), while the DDAC supports continuous action values. The DDAC consisted of two networks; an Actor Network which is a neural network responsible for taking actions based on perceived states and the Critic Network which criticises the value of the action taken. The experimental results showed that the DQN algorithm suffered in performance due to the fact that it cannot support continuous actions or state spaces. The DQN algorithm is suitable for continuous (input) states, however it still requires discrete actions since it finds the action that maximises the action-value function. This would require an iterative process at every time step for continuous action spaces \cite{Lillicrap2015}. As shown in Fig. \ref{fig_dqnvsddac}, the ability to support continuous action values allowed the DDAC algorithm to follow curved tracks more smoothly and stay closer to the centre of the lane when compared to the DQN algorithm, thereby producing better performance for lane keeping.

Vision based vehicle control using CNNs has also been researched. For instance, Zhang et al. \cite{zhang2016query} proposed a supervised learning method, SafeDAgger, for training a CNN to drive in a TORCS simulation. The proposed method is based on the Dataset Aggregation (DAgger) imitation learning algorithm \cite{Ross2011}. In DAgger, the agent first learns a primary policy through traditional supervised learning, with the training set generated by a reference policy. Then, the algorithm iteratively generates new training examples through the learned policies, which are then labelled by the reference policy. The new expanded dataset can then be used to update the learned policy through supervised learning. This has the advantage that states which were not reached in the initial training set can be covered in the new extended training set. The primary policy is then iteratively fine-tuned using the new training set. Zhang et al. proposed an extension to this method, called SafeDAgger, where the system estimates (in any given state) whether the primary policy is likely to deviate from the reference policy. If the primary policy is likely to deviate by more than a specified threshold, the reference policy is used to drive the vehicle instead. The safety policy is estimated by a fully connected network where the input is the last convolutional layer's activation. The authors used this method to train a CNN to predict a continuous steering wheel angle and a binary decision for braking (brake or do not brake). The authors then evaluated supervised learning, DAgger, and SafeDAgger by driving them on three test tracks, with up to three laps on each track. Out of the three algorithms evaluated, SafeDAgger was found to perform best in terms of the number of completed laps, number of collisions, and mean squared error of steering angles. In another work, Pan et al. \cite{pan2018agile} used DAgger-like imitation learning to learn to drive at high speeds autonomously, with continuous actions for both steering and acceleration. The reference policy for the dataset was obtained from a model predictive controller operated using expensive high resolution sensors, which the CNN then learned to imitate using only low cost camera sensors for observations. The technique was first tested in Robot Operation System (ROS) Gazebo \cite{koenig2004design} simulations, followed by a real-world 30m long dirt track with a 1/5-scale vehicle. The sub-scale vehicle successfully learned to drive at speeds up to 7.5m/s around the track. Instead of using direct vision for control, Wang et al. \cite{wang2018deep} demonstrated that DAgger can be used to train an object-centric policy, which uses salient objects in the image (e.g. vehicles, pedestrians) to output a control action. The trained control policy was tested in Grand Theft Auto V simulation, with a discrete control action (left, straight, right, fast, slow stop) which was then translated to a continuous control with a PID controller. The test results demonstrated improved performance with the object-centric policy compared to models without attention or those based on heuristic object selection. Vision based techniques have also been used to mitigate collisions by Porav \& Newman \cite{porav2018imminent}, who built on the previous work by Chae et al. \cite{Chae2017} by using a deep reinforcement learning algorithm for collision mitigation which can provide continuous control actions for both velocity and steering. The system uses a Variational AutoEncoder (VAE) coupled with an RNN to predict the movement of obstacles and learns a control policy with Deep Deterministic Policy Gradient (DDPG) to mitigate collisions in low TTC scenarios. The network used a semantically segmented image to predict continuous steering and deceleration actions. The proposed technique shows improvement over braking-only policies for TTC values between 0.5 and 1.5s, and up to 60\% reduction in collision rates.

Inverse Reinforcement Learning (IRL) approaches have also been investigated in the context of control systems as a way to overcome the difficulty of defining an optimal reward function. IRL is a subset of reinforcement learning, in which the reward function is not specified, but the agent attempts to learn it from an expert's demonstrations. In IRL, the agent assumes that the expert is completing the task by following an unknown reward function. It then estimates a reward function in which the demonstrators' trajectory is the most likely one. This has the advantage that instead of requiring the developer to explicitly specify a reward function, they simply have to demonstrate the intended behaviour. This can be advantageous since in large and complex tasks, defining an adequate reward function to provide optimal agent behaviour can be both difficult and time consuming \cite{Zhifei2012}. IRL approaches have been shown to not only reduce the amount of time required for design and optimisation, but also improve the system performance by creating more robust reward functions. Abbeel \& Ng \cite{Abbeel2004} showed that when IRL was applied to a problem where the agent learned by observing an expert, the agent performed as well as the expert when evaluated with respect to the reward function used by the expert, even if the reward function derived from observations was not the expert's true reward function. Moreover, it was shown that in a simplistic highway driving scenario with 5 different actions for lane selection available to the agent and multiple driving styles demonstrated, the IRL algorithm successfully learned to mimic the demonstrated driving behaviours. Further, Silver et al. \cite{Silver2013} used an IRL algorithm based on Maximum Margin Planning \cite{Ratliff2006} which was shown to be effective in a demonstration of an autonomous vehicle in unstructured terrain. The vehicle was shown to perform better than an agent based on traditional reinforcement learning with a hand-tuned reward function. Additionally, the IRL approach was shown to require significantly less time to design and optimise compared to the reinforcement learning agent. Kuderer et al. \cite{Kuderer2015} proposed a vehicle controller that can learn individual driving styles from demonstration using IRL. The algorithm assumes that the demonstrator is driving in a way to maximise an unknown reward function. From this, the learning model estimates the weights in a linear reward function based on 9 features for driving. Initially, the weights were equally set and were then updated based on demonstrations of 8 minutes per driver. After finding the driving policy, the chosen trajectories were compared to those observed from human drivers. The system was shown to learn drivers' personal driving styles from minimal training data and performed adequately in simulated testing. 

Building on the IRL approaches, Wulfmeier et al. \cite{Wulfmeier2017} proposed an IRL approach for deep learning. The proposed algorithm is based on the Maximum Entropy \cite{Ziebart2008} model for a trajectory planner, and uses CNNs to infer the reward functions from expert demonstration. The approach was trained on a dataset collected over the course of one year with a total of 120km of driving a modified golfcart on walkways and cycle lanes. The input to the network was the LIDAR point cloud map, which was represented on a discretised grid map. The output of the network was a discrete set of actions. The proposed approach was demonstrated to work better than a manually constructed cost function. Moreover, the learned algorithm was shown to be more robust to sensor noise. This shows that the use of DNNs in an IRL algorithm for trajectory planning was beneficial overall. Therefore IRL techniques could be considered as a potential way to overcome the difficulties of designing an optimal reward function for driving.

However, there are some challenges for IRL approaches in practical applications. Firstly, there is no guarantee of optimality of the demonstrations. For example, in a driving demonstration, no human driver can carry out the driving tasks optimally every time. Therefore, the training data will include suboptimal demonstrations which will affect the final reward function constructed. There are some solutions to minimising the effect of suboptimal demonstrations; using multiple trajectories and averaging over multiple sets to find a reward function or removing the assumption of global optimality \cite{Levine2012}. Secondly, reward ambiguity can lead to further problems in IRL approaches. Given expert demonstrations of driving strategies, there can be multiple reward functions that explain the expert's behaviour. Therefore, an effective IRL algorithm must find a reward function that considers the expert's trajectory optimal and rejects other possible trajectories. Thirdly, the reward function derived through IRL methods may not be safe, as noted by Abbeel et al. \cite{Abbeel2007}, who used IRL to operate an autonomous helicopter and had to manually tune the reward function for safety. Therefore, hand tuning of the derived reward function may be required to ensure safe behaviour. Lastly, the computational burden of IRL methods can be heavy since they often require iteratively solving reinforcement learning problems with each new reward function derived \cite{Zhifei2012}. Nevertheless, in tasks where an adequately accurate reward function cannot be easily defined, IRL approaches can provide an effective solution.

While the previously mentioned works in this section demonstrate that a DNN can be trained to drive a vehicle, training a vehicle to simply follow a road or keep in its lane without any outside context is not sufficient for deploying fully autonomous vehicles. Humans drive vehicles with the goal of arriving at our target destination, and learning to drive from camera images to imitate human driving behaviour is not enough to understand the full context behind the human driver's action. For instance, it has been reported \cite{Pomerleau1989}, that upon reaching a fork in the road end-to-end driving techniques tend to oscillate between the two possible driving directions. Not only is this impractical if our goal is to continue in the left direction, but can result in unsafe behaviour where the DNN oscillates between left and right but never picking either direction. Aiming to provide autonomous vehicles with contextual awareness, Hecker et al. \cite{hecker2018end} collected a data set with a 360-degree view from 8 cameras and a driver following a route plan. This data set was then used to train a DNN to predict steering wheel angle and velocities from example images and route plans in the data set. Qualitative testing was done to evaluate learning on instances from the data set, suggesting the model was learning to imitate the human driver, but no live testing was completed to validate performance. With a similar aim, Codevilla et al. \cite{codevilla2018end} trained a supervised learning algorithm, which uses both images and a high-level navigational command for its driving policy. The network was trained through end-to-end supervised learning, conditioned by a high-level command which could be follow road, go straight, turn left, or turn right. The authors tested two network architectures which could take the navigational command into account; one where the command was an additional input to the network, and one where the network branched at the end into multiple sub-modules (feedforward layers), one for each possible command. The authors noted that the latter architecture performed better. The resulting network was initially tested in CARLA \cite{Dosovitskiy17} simulation, followed by real-world testing on a 1/5-scale car. The resulting policy successfully learned to turn the correct way at intersections as commanded. The authors noted that data augmentation and noise injection during training was key to learning a robust control policy. This method was further extended in \cite{codevilla2019exploring}, by using an extra module for velocity prediction, which helps the network in some situations, such as when the vehicle is stopped at a traffic light, to predict the expected vehicle velocity from visual cues and prevent it from getting stuck when the vehicle comes to a full stop. Further improvements to the model were a deeper network architecture and a larger training set, which reduced the variance in training. A slightly different approach was explored using reinforcement learning by Paxton et al.\cite{paxton2017combining} where the high-level command is provided by another DNN responsible for decision making. The system consisted a DDPG network for low-level control and a DQN for a stochastic high level policy subject to linear temporal logic constraints. The aim of the vehicle was to navigate a busy intersection, where some lanes had stopped vehicles so that the host vehicle had to successfully change lanes as well. The system was tested in 100 simulated intersections with and without stopped cars ahead, for a total of 200 tests. Without stopped cars the agent succeeded every time, whereas with stopped cars ahead, 3 collisions occurred.

Moving away from end-to-end approaches, researchers at Waymo recently presented ChauffeurNet \cite{bansal2018chauffeurnet}. ChauffeurNet uses mid-to-mid learning to learn a driving policy, where the input is a pre-processed top-down view of the surrounding environment which represents useful features such as roadmap, traffic lights, a route plan to follow, dynamic objects, and past agent poses. The agent then processes these inputs through an RNN to provide a heading, speed, and waypoint, which are then achieved through a low-level controller. This had the advantage that pre-processed inputs could be obtained either from simulation or real-world data, which makes transferring driving policies from simulation to the real world easier \cite{pan2017virtual, muller2018driving}. Furthermore, synthesising perturbations to model recoveries from incorrect lane positions or even scenarios such as collisions or driving off-road provides the model with robustness to errors and allows the model to learn to avoid such scenarios.

An overview of full vehicle control approaches can be seen in Table \ref{table_sim}. Unlike previous sections, a variety of learning strategies have been utilised here, however supervised learning is still the preferred approach. An important note on the works where full vehicle control via neural networks is researched, is that robust and high performing models still seem out of reach. For instance, techniques which implement full vehicle control tend to have poorer performance on steering than techniques which only consider steering. This is explained by the significant increase in the complexity of the task which the neural network is trained to perform. For this reason, several of the works summarised in this section have been trained and evaluated in simplified simulated environments. While full vehicle control should be the end goal of autonomous vehicle control techniques, current approaches have yet to achieve adequate performance in complex and dynamic environments. Therefore future research is required to further improve the control performance of neural network-driven autonomous vehicles.

\begin{table*}[t]
	\scriptsize
	\renewcommand{\arraystretch}{1.2}
	\caption{A Comparison of Full Vehicle Control Techniques.}
	\label{table_sim}
	\centering
	
	\begin{tabular}{ | m{1.5em} | m{2cm} | m{2cm} | m{2cm} | m{2cm} | m{2cm} | m{2cm} | m{2cm} |}
		\hline
		Ref. & Learning Strategy & Network & Inputs & Outputs & Pros & Cons & Experiments \\
		\hline
		\cite{WeiXia12016} & Supervised Reinforcement Learning & Feedforward network with 2 hidden layers & Not mentioned & Steering, acceleration, braking & Fast training & Unstable (Can steer off the road) & Simulation \\
		\hline
		\cite{Sallab2016} & Reinforcement Learning & Fully connected / Actor-Critic Network with feedforward networks & Position in lane, velocity & Steering, gear, brake, and acceleration values (discretised for DQN) & Continuous policy provides smooth steering & Simple simulation environment & Simulation \\
		\hline
		\cite{zhang2016query} & Supervised Learning & CNN / Feedforward & Simulated camera image & Steering angle, binary braking decision & Estimates safety of the policy in any given state, DAgger provides robustness to compounding errors & Simple simulation environment, simplified longitudinal output & Simulation \\
		\hline
		\cite{pan2018agile} & Supervised Learning & CNN & Camera image & Steering and throttle & High speed driving, Learns to drive on low cost cameras, Robustness of DAgger to compounding errors & Trained only for elliptical race tracks with no other vehicles, Requires iteratively building the dataset with the reference policy & Real world (sub-scale vehicle) \& Simulation \\
		\hline
		\cite{wang2018deep} & Supervised Learning & CNN & Image & 9 discrete actions for motion & Object-centric policy provides attention to important objects & Highly simplified action space & Simulation \\
		\hline
		\cite{porav2018imminent} & Reinforcement Learning & VAE-RNN & Semantically segmented image & Steering, acceleration & Improves collision rates over braking only policies & Only considers imminent collision scenarios & Simulation \\
		\hline 
		\cite{Wulfmeier2017} & Inverse Reinforcement Learning & CNN & LIDAR point clouds on a grid map & Discrete motions & Robust to noise, avoids handcrafting of cost function & Increased computation burden of IRL, no guarantee of cost function optimality & No live testing \\
		\hline
		\cite{hecker2018end} & Supervised Learning & CNN & 360-degree view camera image, route plan & Steering angle, velocity & Takes route plan into account & Lack of live testing & No live testing, tested on data set image examples only \\
		\hline
		\cite{codevilla2018end} & Supervised Learning & CNN & Camera image, navigational command & Steering angle, acceleration & Takes navigational commands into account, generalises to new environments & Occasionally fails to take correct turn on first attempt & Real (sub-scale vehicle) \& Simulation \\
		\hline
		\cite{paxton2017combining} & Reinforcement Learning & Feedforward network with 1 hidden layer & Host vehicle states, set of features for each nearby vehicle, vehicle position and priority in intersection & steering angle rate, acceleration & Considers decision making provided by another DNN & Large number inputs which would be difficult to extract in reality, Not collision free & Simulation \\
		\hline
		\cite{bansal2018chauffeurnet} & Supervised Learning & CNN-RNN & Pre-processed top-down image of surroundings & Heading, velocity, waypoint & Ease of transfer from simulation to real world, robust to deviations from trajectory & Can output waypoints which make turns infeasible, can be over aggressive with other vehicles in new scenarios & Real world \& Simulation \\
		\hline
	\end{tabular}
\end{table*}

\section{Challenges}
The previous section discussed various examples of deep learning applied to vehicle controller design. While this shows that there is a significant amount of interest in the research of such systems, they are still far from ready for commercial application. There remains a number of challenges that must be overcome before learned autonomous vehicle technology is ready for widespread commercial use. This section is dedicated to discussing the technological challenges for deep learning based control of autonomous vehicles. It is worth remembering that besides these technological challenges, issues such as user acceptance, cost efficiency, machine ethics for artificial intelligence technologies, and lack of legislation/regulation for autonomous vehicles must also be addressed. However, the aim of this manuscript is to focus on deep learning based autonomous vehicle control methods and their technical challenges, therefore general and non-technological challenges for autonomous vehicles are out of the scope of this manuscript, for further reading on these topics, see \cite{Maurer2016, EuropeanCommission2016, Bagloee2016, HERETechnologies2017, Bosankic2017, Abraham2017}.

\subsection{Computation}
The major drawback for deep learning methods is the large amount of data and time required for adequate training, especially for reinforcement learning methods. This can lead to long training periods which can cause delays and additional cost in the design of an autonomous vehicle. The common solution to reduce training data requirements or the time required for training is to combine reinforcement learning with supervised learning, which helps reduce the training time whilst still providing good adaptability. Nevertheless, for a fully autonomous vehicle, the amount of training data required to build a reliable and robust system can be vast. It is challenging to train a vehicle to drive in all possible scenarios that it could encounter in the real world due to the huge quantity of data that needs to be collected. There are several companies researching autonomous driving using machine learning and collaborating and sharing data would be the fastest route to move from experimental systems to commercial ones. However, this is unlikely as companies researching autonomous vehicles are not willing to share their resources due to fear of diluting their competitive advantage \cite{Knight2016}. However, while increasing the amount of available data is useful to learn more complex behaviours, using larger data sets brings its own challenges, such as ensuring diversity of the data. If the amount of data used for training the model is increased, without ensuring variety in the data set, the risk of overfitting to the data set increases. For instance, Codevilla et al. \cite{codevilla2019exploring} compared 4 driving models trained with 2, 10, 50, and 100 hours of data, and it was shown that the model trained with 10 hours of driving data performed best in most scenarios. This is due to many of the instances in the training set being very similar, captured in typical driving conditions. As the data set size increases, rare driving scenarios (where the model is more likely to fail) are encountered increasingly rarely during training. Therefore, when generating large data sets, diversity in the data set must be ensured.

Further computational complexity is caused by the continuous states and actions in which the agent has to operate. As stated in the previous section, continuous action values are necessary for a deployable vehicle control system to have adequate performance. However, as the number of dimensions grows, the computational complexity grows exponentially \cite{Kober2013}, this is known as the Curse of Dimensionality \cite{Bellman1966}. In the high-dimensional problems of vehicle control, this has a significant effect on the computational complexity of any solution. Although discretisation of the system can reduce the complexity, as seen in previous examples, this can lead to degradation in system performance. Other solutions include using multiple learners to reduce learning time \cite{Gu2016, Barron2009}, evolution strategies which are highly parallelisable \cite{Salimans2017}, or removing unnecessary data from the training and system input data \cite{Yang2017}.

Overall, the high computational burden of DNNs is a challenge to not only the development and training of the networks but also the deployment of such systems in vehicles. The high computational overhead of the deep learning algorithms will require high computing capabilities on-board, driving up the system cost and power requirements, which must be kept in mind during the system design.

\subsection{Architectures}
Another challenge with deep learning is selecting the architecture of the neural networks. There are no clear guidelines for 'good' neural network architecture for a given task. For instance, in terms of size and number of layers, it has been shown that too few neurons will lead to a system with poor performance. However, too many neurons may overfit to the training data and therefore not generalise well. Also, given that additional neurons will lead to increased computational complexity, finding an optimal number of neurons would be of great benefit to deep learning methods \cite{LeCun1989, Morgan1989}. Other parameters can also have an effect on the performance, training, and convergence of the system. The fundamental architecture, training method, learning rate, loss function, batch size etc., all need to be decided upon and defined, which affect the performance of the agent. However, there are few methods for choosing these parameters, and often trial-and-error and heuristics are the only viable options for optimising each parameter due to the complexity of DNNs \cite{Nielsen2015}. This is generally achieved by choosing a range of values for the hyperparameters in the neural network, and finding the best performing values. However, using such trial-and-error methods for exploring the hyperparameter space can be slow, given the amount of computation required for each training run.

Solutions to this challenge currently being researched include computerised ways of finding optimal values for these parameters, either by trialling across a range or using model-based methods to converge on the best values. There are several methods for changing the parameters over the chosen range, such as Coordinate Descent \cite{bengio2012practical}, Grid Search \cite{bengio2012practical, lecun1998efficient}, and Random Search \cite{BergstraJAMESBERGSTRA2012}. Coordinate Descent keeps all hyperparameters except one fixed, and finds the best value for one parameter at a time. Grid Search optimises every parameter simultaneously, including the cross-product of all intervals. However, this vastly increases the computational expense by requiring a large number of neural network models to be trained and therefore is only suitable when the models can be trained quickly. Random Search often finds a good set of parameters faster than a Grid Search by sampling the chosen interval randomly \cite{BergstraJAMESBERGSTRA2012, Raffel2015}. However, this has the disadvantage that the parameter space is often not covered completely, and some sample points can be very close to each other. These disadvantages can be solved by using quasi-random sequences \cite{Sevchuk2016}. Alternatively, one can use model-based hyperparameter optimisation methods, such as Bayesian optimisation or tree-structured Parzen estimators, which tend to yield better results but are more time intensive \cite{Sevchuk2016, Snoek2012, Rasmussen2004, Bergstra2011}. Other proposed approaches focus on automated hyperparameter tuning by eliminating undesirable regions of the hyperparameter search space in order to converge to optimal values \cite{kumar2018parallel, hashimoto2018derivative}. Recent research has also explored neural architecture search methods which take hardware efficiency into account by incorporating the hardware feedback into the learning signal \cite{cai2019once, tan2019mnasnet, wu2019fbnet, scheidegger2019constrained}. This has resulted in neural network architectures which are specialised for specific hardware platforms, and demonstrate a hardware efficiency benefit over non-specialised architectures. Such methods could also be extended to find efficient network architectures for vehicle on-board hardware platforms. It should be noted that automated neural architecture search is an active area of research, for further discussion on this topic we refer the reader to the survey by Elsken et al. \cite{elsken2019neural}.

While architecture selection is a general problem for many deep learning applications, a complex task such as autonomous driving also brings its own challenges. Currently, most end-to-end driving systems have been limited to smaller networks. This is due to the relatively small datasets used, which would cause deeper networks to overfit to the training data. However, as noted in \cite{codevilla2019exploring}, when large amounts of data are available, deeper architectures can reduce both bias and variance in training, resulting in more robust control policies. Further thought should be given to architectures specifically designed for autonomous driving, such as the conditional imitation learning model \cite{codevilla2018end}, where the network included a different final network layer for each high-level command used for driving. These challenges translate to mid-to-mid approaches as well, as the selection of high-level features represented in the input to the network must be chosen carefully. Future works investigating specialised network architectures for autonomous driving can therefore be expected.

\subsection{Goal Specification}
Adequate goal specification is a challenge specific to reinforcement learning methods. One of the advantages of reinforcement learning is that the behaviour of the agent does not need to be specified implicitly as it would be in rule-based systems. Only the reward function, which can often be easier to define than the value function, and the control action (e.g. steering, acceleration, braking) need to be defined. However, the goal of reinforcement learning is to maximise the long term accumulated reward as defined by the reward function. Therefore, the desired behaviour of the agent must be accurately captured by the reward function, otherwise unexpected and undesired behaviour might occur. For instance, instead of using binary rewards for successful or unsuccessful completion of tasks, intermediate rewards can be used to guide the agent towards desired behaviour, this process in known as reward shaping \cite{Ng1999, Laud2004}. For example, Desjardins \& Chaib-Draa \cite{Desjardins2011} used the time headway derivative to reward the agent for actions that helped it move towards the ideal time headway state. Furthermore, for a complex task such as driving, a multi-objective reward function needs to consider different objectives which may conflict with each other. For example, for driving, these objectives may include maintaining a safe distance from other vehicles, staying close to the centre of the lane, avoiding pedestrians, not changing lanes too often, maintaining desired velocity, and avoiding harsh accelerations/braking. Hence, the reward function should not only consider all factors that affect the agent's behaviour, but also the weight of these factors. 

A further challenge for agents which control both lateral \& longitudinal actions is the difficulty of defining a reward function when the agent must be able to perform multiple actions (steering, braking, and acceleration). In reinforcement learning, the agent uses the feedback from the reward function to improve its own performance. However, when the agent is carrying out multiple actions, it may not be clear which of the actions resulted in the given reward. For example, if the vehicle steers away from the road, the acceleration may not be at fault but a negative reward signal is sent to the agent. One solution to this is a Hybrid Reward Architecture \cite{VanSeijenHarmFatemiMehdiRomoffJoshuaLarocheRomainBarnesTavianTsangJeffrey2017}, where the system uses a decomposed reward function and learns a separate value function for each component reward function. Alternatively, Shalev-Shwartz et al. \cite{Shalev-Shwartz2016a} proposed a solution in which the reward function is decomposed into a high level decision making system, through which the agent learns to drive safely and make strategic decisions (e.g. which cars to overtake or give way to), and a low level reward function which helps the agent learn an optimal policy for different actions (e.g. overtaking, merging, decelerating etc.). 

The developer should also take care that the agent does not exploit the reward function in unexpected ways, resulting in unintended behaviour. This effect is also known as reward hacking. Reward hacking occurs when the agent finds an unanticipated way of exploiting the reward function to gain large rewards in a way which goes against the developers' defined objective(s) for the agent.  For example, a robot used in ball paddling with a reward function based on the distance between the ball and the desired highest point, may attempt to move the racket up and keep the ball resting on it \cite{Kober2013}. Potential solutions to avoid reward hacking were proposed by Amodei, et al. \cite{Amodei2016} in the form of adversarial reward functions, model look-ahead, reward capping, multiple reward functions, and trip wires. Adversarial reward functions utilise a reward function which is its own agent, similar to generative adversarial networks. The reward function agent can then explore the environment, making it more robust to reward hacking. It could, for example, try to find instances where the system claims a high reward from its actions, while a human would label it as a low reward. On the other hand, model look-ahead gives a reward based on anticipated future states, instead of the present one. Reward capping is a simple solution to reward hacking, where a maximum value is imposed on the reward function, thereby preventing unexpected high reward scenarios. Multiple reward functions can also increase robustness to reward hacking, since multiple rewards can be more difficult to hack than a single one. Finally, trip wires are deliberately placed vulnerabilities in the system, where reward hacking is most likely to occur. These vulnerabilities are then monitored to alert the system if the agent is attempting to exploit its reward function. Another approach to solving these challenges in goal specification is using inverse reinforcement learning to extract a reward function from expert demonstrations of the task \cite{Russell1998, Kolter2008, Silver2010, Ratliff2008}.

\subsection{Adaptability \& Generalisation}
Another challenge for learned control systems is dealing with different environments with a scalable approach. For example, a driving strategy that is successful in an urban environment may not be optimal on a highway, since they are very different environments with different traffic flow patterns and safety issues. Similar issues arise with changing weather conditions, seasons, climates etc. A neural network's ability to use what it has learned from previous experiences to operate in a completely new environment is referred to as generalisation. However, the problem with generalisation is that even if the system demonstrates good generalisation in one new environment, there is no guarantee it will generalise to other possible environments. Moreover, considering the complex operating environment of a vehicle, it is not possible to test the system in all scenarios. Therefore, building a deep learning system capable of generalising to such a vast variety of situations, as well as validating its generalisation capability poses major challenges. This is a challenge that must be overcome for deep learning driven autonomous vehicles to be deployable in the real world, as the vehicles must be able to cope with the various different environments it will be used in.

Generally, to avoid poor generalisation in DNNs the training must be stopped before the DNN starts to overfit to the training data. Overfitting refers to creating a model that fits the training data too well, losing its ability to generalise to new data. Overfitting occurs when the network is trained with either insufficient amounts of training data or too many training episodes on the same training data. This results in the neural network memorising the training data, thereby losing generalisation. Unfortunately, there are no known methods of choosing the optimal stopping point in order to avoid overfitting \cite{Schalkoff1997}. However, it is possible to get some indication of the network's generalisation capability by having three different data sets: training, validation, and test sets. The training and validation sets are used during training, but only the training set results are used to update the network weights \cite{Ripley1996}. The purpose of the validation set is to minimise overfitting, by monitoring the error in the validation data set. In this way, it will be ensured that changes which reduce the error on the training set also reduce the error on the validation set, thereby avoiding overfitting. If the accuracy of the validation set starts to decrease over the training iterations, then the network is starting to overfit and training should be stopped. In addition to stopping overfitting, a validation set can also be used to compare different network architectures (e.g. comparing two different networks with different numbers of hidden layers) to provide a measure of generalisation. Nevertheless, utilising the validation set simultaneously in the selection of the network and to terminate training can result in overfitting to the validation set. Therefore an additional independent set, known as testing set, is required for the evaluation of the network performance \cite{Bishop1995}. The testing set is only used to test the final network to confirm its performance and generalisation capabilities. The testing set must provide an unbiased evaluation of the network's generalisation \cite{Ripley1996}. Therefore, it is crucial that the test set is not used to choose between different networks or network architectures.

There are also techniques available for DNNs which aim to reduce the test error, although often at the cost of increased training error, known as regularisation techniques \cite{GoodfellowIanBengioYoshuaCourville2016}. The basis of regularisation techniques is to introduce some constraints on the deep learning model, which either introduce prior knowledge into the model or promote simpler models in order to achieve better generalisation capability. There are a variety of regularisation techniques available to choose from. For instance, L1 and L2 regularisation techniques introduce a constraint on the model by including an additional term in the cost function of the learning model, which makes the network prefer smaller weights. The smaller weights in the network reduce the effect of individual inputs on its behaviour, which means that the effect of local noise is reduced and the network is more likely to learn trends across the whole data set \cite{Nielsen2015, ng2004feature}. Similarly, imposing constraints on the network weights through weight clipping has also been shown to improve robustness \cite{merolla2016deep, courbariaux2015binaryconnect}. Another popular regularisation technique is dropout, which drops out some randomly selected neurons from training and only updates the remaining weights for the given training example. At each weight update, a different set of neurons is omitted, thereby preventing complex co-adaptions between neurons. This helps each neuron learn features which are important for the given task and therefore helps reduce overfitting \cite{HintonGeoffreyE.SrivastavaNitishKrizhevskyAlexSutskeverIlyaSalakhutdinov2012, Srivastava2014}.

\subsection{Verification \& Validation}
The testing of the system needs to be rigorous to validate the performance and safety of the system. However, the problem is that real-world testing can be expensive in terms of time, labour, and finances. Indeed, full-scale vehicle studies with multiple vehicles have typically been achieved through collaboration of government research projects with automotive manufacturers, such as Demo '97 \cite{Raza1996, Rajamani2000, Thorpe1997} or Demo 2000 \cite{Kato2002}. Alternatively, simulation studies can reduce the amount field testing required, and can be used as a first step for performance and safety evaluation. Simulation studies are significantly cheaper, faster, more flexible, and can be used to set up situations not easily achieved in real life (e.g. crashes). Indeed, with the increasing accuracy and speed of simulation tools, simulation has become an increasingly dominant method of study in this field \cite{Ng2008b}. 

While simulation has multiple advantages, the model errors must be kept in consideration throughout the verification and validation process. This is especially critical for training, as training an agent in an imprecise model will result in a system that will not transfer to the real-world without significant modifications \cite{Kober2013, Atkeson1994}. Complex mechanical interactions, such as contacts and friction, are often difficult to model accurately. These small variations between the simulation model and the real-world can have drastic consequences on the system behaviour in the real world. In other words, the problem is the agent overfitting its policies to the simulation environment, and not transferring well to a real-world environment. For a system that can be evaluated and used in the real-world, training, as well as testing, in both simulation and field tests would be required \cite{hester2017learning}. The large number of trials required for reinforcement learning algorithms to converge, makes them susceptible to this issue where simulation is used for training. However, recent studies in robot manipulation have shown effective transfer of learned policies from simulation to the real world \cite{christiano2016transfer, rusu2016sim, tzeng2015towards, tobin2017domain}.

Validation of the model and simulation environment alone is not enough for autonomous vehicles, as the influence of the training data can be equal to that of the algorithm itself \cite{varshney2017safety}. Therefore, there should be emphasis on validating the quality of the training set as well. Ensuring that the data set represents the desired operational environment adequately, and covers the potential states is important. For instance, data sets that are biased towards a certain action (e.g. turn left) or scenario (e.g. driving in daytime) can introduce harmful biases into the learning model. Therefore, data sets should be validated to understand if they contain potentially harmful biases or patterns that could lead to undesirable behaviour of the learned control policy \cite{torralba2011unbiased}.

\begin{table*}[t]
	\scriptsize
	\renewcommand{\arraystretch}{1.2}
	\caption{A Summary of Research Challenges.}
	\label{table_challenges}
	\centering

	\begin{tabular}{ | m{1.5cm} | m{7.5cm} | m{7.5cm} |}
		\hline
		Challenge & Sub-challenges & Potential Solutions \\
		\hline
		Computation & \begin{itemize}
			\item Computation requirements for deep learning
			\item Large data sets for supervised learning
			\item Curse of dimensionality in high dimensional problems
			\item Simulation requirements for sample inefficient techniques
		\end{itemize} & \begin{itemize}
			\item Scalable model architectures
			\item Sharing data sets
			\item Improving sample efficiency in reinforcement learning
			\item Parallelisable architectures for training
		\end{itemize} \\
		\hline
		Architectures & \begin{itemize}
			\item Lack of clear rules for network architectures
			\item Reliance on heuristics and trial-and-error
		\end{itemize} & \begin{itemize}
			\item Automated neural architecture search methods
			\item Specialised architectures for autonomous driving
		\end{itemize} \\
		\hline
		Goal Specification & \begin{itemize}
			\item Well designed reward functions for complex tasks
			\item Multi-objective reward functions
			\item Reward hacking
		\end{itemize} & \begin{itemize}
			\item Reward shaping
			\item Inverse reinforcement learning
			\item Hybrid reward architectures
		\end{itemize}\\
		\hline
		Adaptability \& Generalisation & \begin{itemize}
			\item Wide variety of the operational environment
			\item Overfitting to training data/environment
		\end{itemize} & \begin{itemize}
			\item Representative data sets and/or training environments
			\item Effective use of regularisation techniques
		\end{itemize}\\
		\hline
		Verification \& Validation & \begin{itemize}
			\item Inability to test in all possible scenarios
			\item High cost of field testing
			\item Inaccuracies in simulation
			\item Biases and gaps in data sets
		\end{itemize} & \begin{itemize}
			\item High fidelity simulations
			\item Effective simulation to real world transfer
			\item Validation of data set coverage
		\end{itemize} \\
		\hline
		Safety & \begin{itemize}
			\item Complexity and opaqueness of DNNs
			\item Safe training in the real world
			\item Adversarial attacks
		\end{itemize} & \begin{itemize}
			\item Research into interpretability of DNNs
			\item Fail safes and virtual safety cages
			\item Human oversight
			\item Improving model robustness to perturbations
		\end{itemize}\\
		\hline
	\end{tabular}
\end{table*}

\subsection{Safety}
In a safety-critical system, such as vehicle operation, a serious malfunction or failure could result in death or serious harm to people or property. Therefore, the safety of road users must be ensured before such systems are deployed commercially. However, ensuring functional safety in deep learning systems can be challenging. As the neural networks become more complex, the solutions they provide and how they come to those solutions becomes increasingly difficult to interpret \cite{castelvecchi2016can}. This is known as the black box problem. The opacity of these solutions is an obstacle to their implementation in safety-critical applications; while it is possible to show that these systems provide good performance in our validation environment, it is impossible to test these systems in all the possible environments they would encounter in the real world. Therefore, if we do not understand the way in which the system makes its decisions, ensuring it does not make unsafe decisions in new environments becomes increasingly difficult. It becomes even more challenging in online learning methods, since they change their policies during operation and therefore could potentially shift from safe policies to unsafe policies over time \cite{zhang2015controller, clark2013study, jacklin2005development, wilkinson2013final, van2017challenges}.

Any autonomous vehicle system not only needs to drive safely, but it also needs to be capable of reacting in a safe manner to other vehicles or pedestrians acting unpredictably. It can be difficult to guarantee the safety for any vehicle controller if, for example, another driver is acting recklessly or a previously unseen pedestrian runs onto the road. Therefore, it would be useful to include unsafe and aggressive driving behaviours of other vehicles into the training data of the vehicle controller to enable it to learn how to deal with such situations. One option to improve reliability and safety in such situations is utilising a trauma memory \cite{Chae2017} where rare negative events (e.g. collisions) are stored. These are then used in training to persistently remind the agent of these events and ensure it maintains safe behaviour.

Also, safety must be maintained during any training or testing in the real world. For instance, during early training of a reinforcement learning agent, the agent is more likely to use exploration than exploitation of past experiences, which means the agent will effectively be learning through trial and error. Therefore, care must be taken to ensure the exploration happens in a safe manner. This is especially true in any environment including other road users or pedestrians, since inappropriate actions chosen due to exploration could have disastrous results. Exploration poses safety challenges as the agent is encouraged to take random actions, which can lead to catastrophic events if not considered beforehand \cite{schneider1997exploiting, bagnell2004learning, deisenroth2011pilco, moldovan2012safe}. Potential solutions include the use of demonstrations such as in IRL to provide examples of safe behaviour which could be used as a baseline policy, simulated exploration where exploration happens in a simulated environment, bounded exploration which limits exploration in state spaces which are considered unsafe, and human oversight although this is limited in scalability and not feasible in some real-time systems. The same holds true for any testing and evaluation of the system; until the system has been deemed to perform adequately and in a safe manner, all necessary precautions must be taken to ensure safety \cite{Amodei2016}.

An approach for ensuring functional safety for deep learning based autonomous vehicles is suggested by Shalev-Shwartz et al. \cite{Shalev-Shwartz2016a}. In the proposed system architecture, the policy function is decomposed into a learnable part and a non-learnable part. The learnable part is responsible for the comfort of driving and for making strategic decisions (e.g. which cars to overtake or give way to). This policy is learned from experience by maximising an expected reward from the reward function. On the other hand, the non-learnable policy is responsible for the safety by minimising a cost function with hard constraints (e.g. the vehicle is not allowed within a specified distance of other vehicles' trajectories) to ensure functional safety. Alternatively, Xiong et al. \cite{Wang2016} suggested a control structure which combines reinforcement learning based control with safety based control and path tracking. The aim is to combine a traditional control method with a reinforcement method to take advantage of the superior performance of deep learning systems whilst ensuring safety through traditional control theory. The path tracking element is included to ensure the vehicle stays on (or as close as is safe to) the centre of the lane. The reinforcement learning approach is based on the DDPG algorithm. Also, the safety based controller uses an Artificial Potential Field method \cite{glaser2010maneuver} which models any obstacles with a repulsive force to steer the vehicle away from them. The final steering policy is then found by the weighted summation of the three models. The system was shown to keep a safe distance in a simulated environment where the vehicle had to drive along a curve with other vehicles nearby.

Furthermore, malicious inputs to deep learning systems have to be considered. It has been shown that visual classification DNN systems are vulnerable to adversarial examples, which are perturbed images that cause the DNNs to misclassify them with high confidence \cite{Szegedy2013, Nguyen2015, Moosavi-Dezfooli2015, Goodfellow2014}, including misclassification of traffic signs \cite{Huang2017a}. DNNs have been shown to be vulnerable to printed adversarial examples in the real world \cite{Kurakin2016} and even to 3D-printed physical adversarial examples \cite{Athalye2017}, which suggests they are a threat to DNN applications in the real world. Moreover, the image modification of the adversarial examples have been shown to be subtle enough that a human eye does not notice the modification, making prevention of such malicious attacks difficult \cite{Kurakin2016}. These types of weaknesses in DNNs could be exploited and pose a security concern for any technology using DNNs. Although defences against these attacks have been proposed \cite{yuan2019adversarial}, state-of-the-art attacks can by-pass defences and detection mechanisms.

\section{Concluding Remarks}
In this manuscript, a survey of autonomous vehicle control approaches utilising deep learning was presented. The approaches were separated into three categories: lateral (steering), longitudinal (acceleration and braking), and simultaneous lateral and longitudinal control methods. The focus of this manuscript has been on the vehicle control techniques rather than perception, however there is some obvious overlap between them. It was shown that research interest in this field has grown significantly in recent years and is expected to continue to do so. The applications discussed in this paper show great promise for the application of deep learning to autonomous vehicle control. However, current deep learning based controller performance has significant room for improvement. Moreover, much of the current research is only limited to simulation. While testing in simulation is useful for feasibility studies and initial performance evaluations, extensive testing and training in the field will be required before these systems are ready for deployment.

The main research challenges to deep learning based vehicle control were also discussed and can be seen summarised in Table \ref{table_challenges}. Computation was identified as a challenge due to the large amount of data required to train deep learning models. Architectures were also identified as a challenge due to the difficulty of choosing the optimal network architecture for a given task. Goal specification is a challenge for reinforcement learning techniques due to the importance of designing a reward function which promotes the desired behaviour. Adaptability and generalisation is a challenge in the autonomous vehicle domain due to the highly complex nature of the operational environment. Verification and validation is a further challenge due to the high cost and time requirements of field tests and training. While simulation is an obvious solution to reduce the amount of physical field testing required, the use of simulation in training and testing has its own drawbacks. Safety was identified as a crucial challenge due to the safety critical nature of the autonomous vehicle domain. This is made more challenging due to the opaque nature of deep learning methods, making safety validation of these systems problematic. 

Therefore, further research into interpretability of neural networks and functional safety validation methods for neural network-driven vehicles will be required. Before deep learning can be deployed on the road, some safety validation techniques will need to be found to address their opaqueness. Ensuring the safety of these deep neural networks is a major barrier preventing them from being used commercially. Furthermore, as noted by Salay et al. \cite{salay2017analysis} in their analysis of the ISO26262 \cite{iso26262} standard, more than 40\% of the required software techniques in the current version of the standard are incompatible with machine learning techniques, whilst the rest are either directly applicable or applicable if modified slightly. This reveals further need for these standards to be revised to address machine learning systems for autonomous vehicles \cite{falcini2017deep}. Other safety aspects which warrant further research include defences against adversarial attacks, as they currently present a significant safety problem for the use of DNNs in autonomous vehicles. Also, robustness to erroneous inputs from sensory data or communication failures must be investigated. There is currently a significant gap in the literature for investigation of fault tolerant systems. Further research into how deep learning control systems deal with issues such as communication failures, erroneous sensory inputs, input noise, or sensor failure would move the industry towards robust and safe solutions. Furthermore, while research into deep neural networks with both lateral and longitudinal vehicle control is still relatively sparse, there is significant on-going research in this area. Full vehicle control with deep neural networks is typically achieved in simple simulation scenarios and/or with discretised outputs. Much work can be done to improve the performance of the full vehicle control techniques as well. Techniques in Sections III-A and B show promising results for lateral and longitudinal control systems, and future work will be required to bridge these techniques into an autonomous vehicle system with strong performance in the more general case of combined lateral and longitudinal control. This will also include further experiments in the real world to validate the performance of the learned control policies. Other avenues for future research include learning driving manoeuvrers which are still typically achieved through classical control techniques, such as overtaking \cite{dixit2018btrajectory, dixit2019trajectory} or merging \cite{amezquita2019experimental, milanes2010automated}. Further work will also be needed to design autonomous vehicles which can understand the rules of the road and the behaviour of other road users. Some on-going research was discussed where the deep neural network can take into account the intended route or target destination, but more research is needed to ensure these techniques can stop at stop signs and red lights, respect speed limits, or negotiate intersections and roundabouts with other vehicles.

\bibliographystyle{IEEEtran}
\bibliography{DL_Survey}

\begin{IEEEbiography}[{\includegraphics[width=1in,height=1.25in,clip,keepaspectratio]{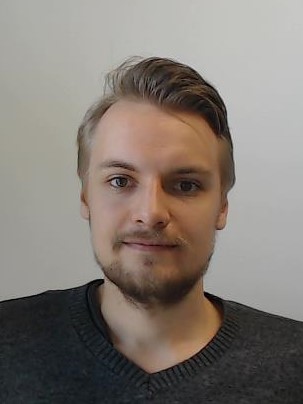}}]{Sampo Kuutti}
	received the MEng degree in mechanical engineering in 2017 from University of Surrey, Guildford, U.K., where he is currently pursuing the PhD degree in automotive engineering with the Connected Autonomous Vehicles Lab within the Centre for Automotive Engineering. His research interests include deep learning applied to autonomous vehicles, functional safety validation, and safety and interpretability in machine learning systems.
\end{IEEEbiography}
\begin{IEEEbiography}[{\includegraphics[width=1in,height=1.25in,clip,keepaspectratio]{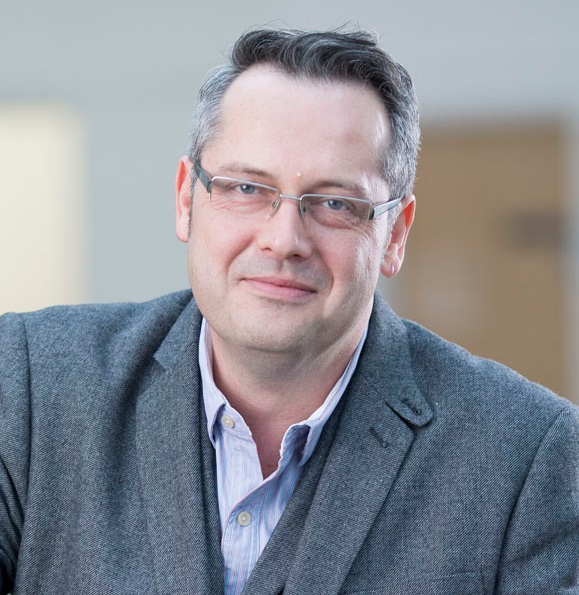}}]{Richard Bowden}
	is Professor of computer vision and machine learning at the University of Surrey where he leads the Cognitive Vision Group within the Centre for Vision, Speech and Signal Processing. His research centres on the use of computer vision to locate, track, and understand humans. He is an associate editor for the journals Image and Vision computing and IEEE TPAMI. In 2013 he was awarded a Royal Society Leverhulme Trust Senior Research Fellowship and is a fellow of the Higher Education Academy, a senior member of the IEEE and Fellow of the International Association of Pattern Recognition (IAPR).
\end{IEEEbiography}
\begin{IEEEbiography}[{\includegraphics[width=1in,height=1.25in,clip,keepaspectratio]{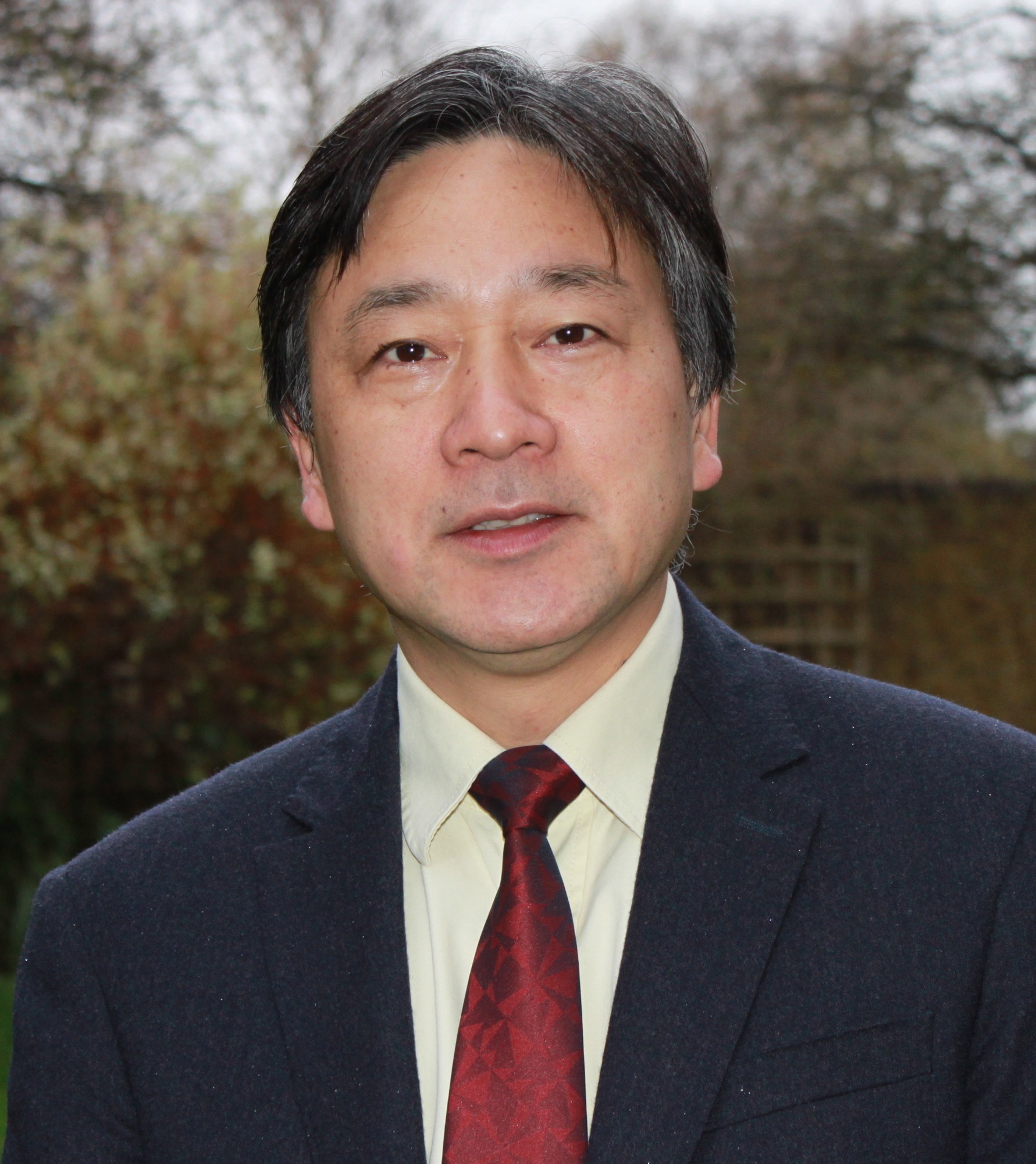}}]{Yaochu Jin}
	is a Professor in Computational Intelligence, Department of Computer Science, University of Surrey, Guildford, U.K. His main research interests include data-driven surrogate-assisted evolutionary optimization, evolutionary learning, interpretable and secure machine learning, and evolutionary developmental systems. 
	
	Dr Jin is the Editor-in-Chief of the IEEE TRANSACTIONS ON COGNITIVE AND DEVELOPMENTAL SYSTEMS and Co-Editor-in-Chief of Complex \& Intelligent Systems. He is an IEEE Distinguished Lecturer and IEEE Fellow.
\end{IEEEbiography}
\begin{IEEEbiography}[{\includegraphics[width=1in,height=1.25in,clip,keepaspectratio]{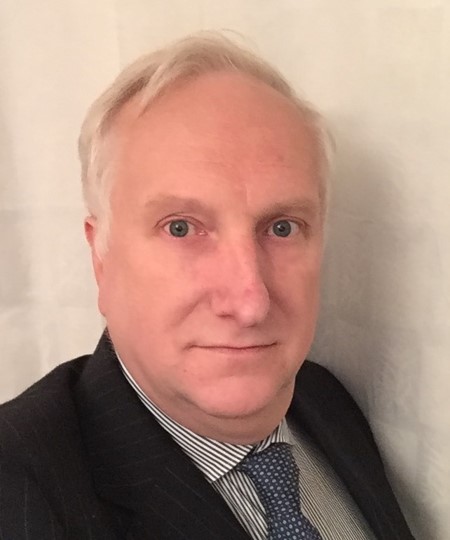}}]{Phil Barber}
	was formerly Principal Technical Specialist in Capability Research at Jaguar Land Rover.
	For over 30 years in the automotive industry he has witnessed the introduction of computer controlled by-wire technology and been part of the debate over the safety issues involved in the implementation of real-time vehicle control. 
\end{IEEEbiography}
\begin{IEEEbiography}[{\includegraphics[width=1in,height=1.25in,clip,keepaspectratio]{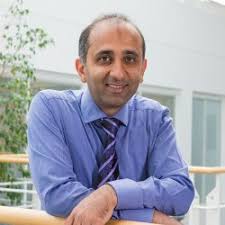}}]{Saber Fallah}is Senior Lecturer (Associate Professor) in Vehicle and Mechatronic Systems at the University of Surrey and the Director of Connected Autonomous Vehicle Lab within the Centre for Automotive Engineering, where he leads several research activities funded by the UK and European governments (e.g. EPSRC, Innovate UK, H2020) in collaboration with major companies active in autonomous vehicle technologies. His research interests include reinforced deep learning, advanced control, optimisation and estimation and their applications to connected autonomous vehicles.
\end{IEEEbiography}

\vfill
\end{document}